\documentclass[letterpaper, 10 pt, conference]{ieeeconf}  

\usepackage{amsmath,graphicx}
\usepackage[T1]{fontenc}
\usepackage{multirow}
\usepackage{floatrow}
\IEEEoverridecommandlockouts                              

\title{\LARGE \bf
CDFI: Cross Domain 
Feature Interaction for Robust Bronchi\\ Lumen Detection
}

\author{Jiasheng Xu$^{1,2}$, Tianyi Zhang$^{1,2}$, Yangqian Wu$^{1,2}$, Jie Yang$^{1,2}$, Guang-Zhong Yang$^{1}$, Yun Gu$^{1,2,3}$
\thanks{This work was supported by National Key RD Program of China (No.
2019YFB1311503), NSFC China (No. 62003208);Committee of Science
and Technology, Shanghai, China (No.19510711200);Shanghai Sailing Program (20YF1420800), and Shanghai Municipal of Science and Technology
Project (Grant No.20JC1419500).}
\thanks{$^{1}$Jiasheng Xu, Tianyi Zhang, Yangqian Wu, Jie Yang, Guang-Zhong Yang and Yun Gu are with 
        the Institute of Medical Robotics, Shanghai Jiao Tong University,
        Shanghai, CHINA. {\tt\small \{xujiasheng, autozty, wyq19981114, jieyang, gzyang, geron762\}@sjtu.edu.cn}}%
\thanks{$^{2}$Jiasheng Xu, Tianyi Zhang, Yangqian Wu, Jie Yang and Yun Gu are also with 
the Institute of Image Processing and Pattern Recognition, Shanghai Jiao Tong University,
Shanghai, CHINA.}%
\thanks{$^{3}$Yun Gu is also with the Shanghai Center for Brain Science and Brain-Inspired Technology, Shanghai, China}%
}

\begin{document}
\maketitle
\begin{abstract}

  Endobronchial intervention is increasingly used as a minimally invasive means for the treatment of pulmonary diseases.
  In order to reduce the difficulty of manipulation in complex airway networks, robust lumen detection is essential for intraoperative guidance. However, these methods are sensitive to visual artifacts which are inevitable during the surgery.
  In this work, a cross domain feature interaction (CDFI) network is proposed to extract the structural features of lumens, as well as to provide artifact cues to 
  characterize the visual features. To effectively extract the structural and artifact features, the Quadruple Feature Constraints (QFC) module is designed to constrain the intrinsic connections of samples with various imaging-quality. Furthermore, we design a Guided Feature Fusion (GFF) module to supervise 
  the model for adaptive feature fusion based on different types of artifacts.  Results show that the features extracted by the proposed method can preserve 
  the structural information of lumen in the presence of large visual variations, bringing much-improved lumen detection accuracy. 
\end{abstract}

\begin{keywords}
Lumen detection, Artifacts, Domain adaption, Feature fusion
\end{keywords}

\section{Introduction}\label{sec:intro}

Recent technical advances in bronchoscopes have improved the endobronchial procedures
in respiratory medicine. With flexible continuum endoscopes and high-resolution imaging,
operators can access the distal small bronchi for real-time precision biopsy or locally focused energy treatment. 
Thus far, the bronchoscopy biopsy has been used in the treatment of 
chronic obstructive pulmonary disease (COPD)~\cite{wan2006bronchoscopic}, 
peripheral nodules~\cite{memoli2012meta} and recent COVID-19 cases~\cite{luo2020performing}.

\begin{figure}[!t]
  \centering
  \centerline{\includegraphics[width=1.0\linewidth]{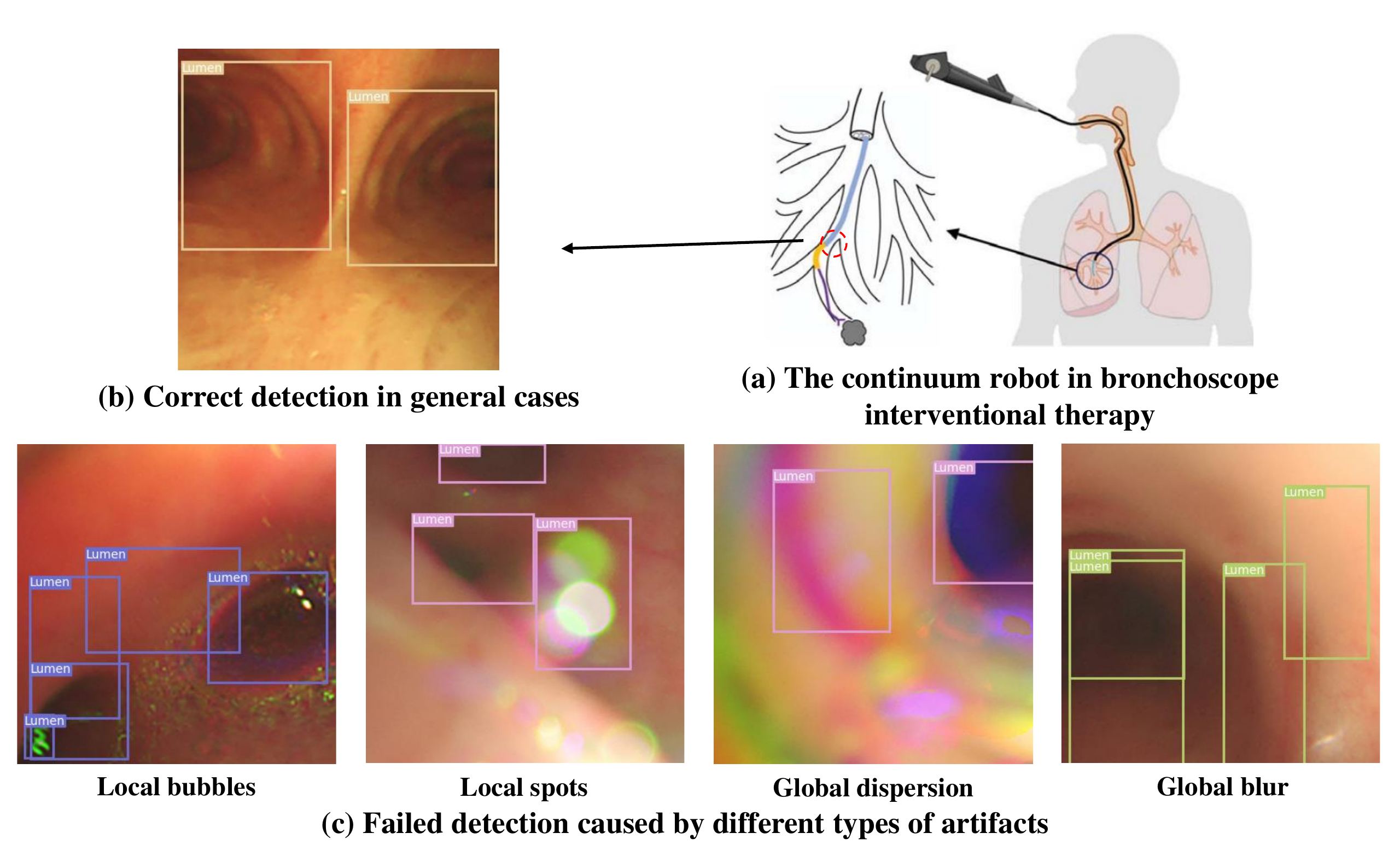}}
  \caption{The continuum guiding robot in bronchoscopic intervention and the lumen detection cases. (a) illustrates the continuum robot's navigation in bronchi~\cite{liu2018design}. (b) shows the correct detection results in general cases that can guide the robot. (c) shows the failed detection results caused by different types of artifacts, local bubbles and spots, global dispersion and blur, which affect the navigation.}\label{fig::robot}
  \end{figure}

Due to the restricted field of view, it remains a challenging task for surgeons 
to manipulate the flexible bronchoscope inside the complex airway networks. 
As an indispensable task in vision-guided navigation, accurate detection and 
localization of lumens can provide the guidance for both surgeons and 
robot-assisted systems~\cite{zou2022robotic}. As shown in Fig. \ref{fig::robot}, 
the localization of lumen centers can guide the forwarding direction for robot-assisted 
bronchoscope during the biopsy. To perform accurate lumen detection, the vision-based methods have
been proposed in literatures. Lim et al.~\cite{lim2006efficient} used Otsu's method to detect the lumens
for automated endoscopic procedures; Wang et al.~\cite{sanchez2013line} proposed a lumen 
detection and segmentation framework based on splitting-merging method; Besides the conventional
techniques, recent works also introduced the deep neural networks for accurate lumen detection.
Two-stage object detection methods~\cite{girshick2014rich,he2017mask,sun2021sparse} can
 achieve promising performance in detection tasks, while single-stage methods~\cite{redmon2018yolov3,liu2016ssd,lin2017focal} 
 balanced the inference speed and detection accuracy. Among them, Huang et al.~\cite{huang2020deep} proposed
 a residual network for lumen contour segmentation; Yen et al.~\cite{yen2021automatic} adopted the
 YOLOv3 model to detect the lumens for the automated orientation of capsule colonoscopes;
 Zou~\cite{zou2022robotic} integrated the single-shot-detection framework (SSD) with the morphological
 operations for lumen detection. Although the methods above have achieved promising accuracy in 
 lumen detection, most of them are tested with phantoms or high-quality endoscopic videos. 
 As shown in Fig. \ref{fig::robot}(b) and \ref{fig::robot}(c), the quality of endoscopic images can be affected
 by visual artifacts due to the patient-specific variations of lumen walls, the instability of robot
 motion and the illumination changes. The detectors trained with clean samples cannot generalize well to the cases 
  with artifacts for lumen detection.  In addition, the uncertainty of visual
variation also adds difficulty to the  learning of lumen structural knowledge from endoscopic images with artifacts.

  \begin{figure*}[!t]
    \centering
    \centerline{\includegraphics[width=1.0\linewidth]{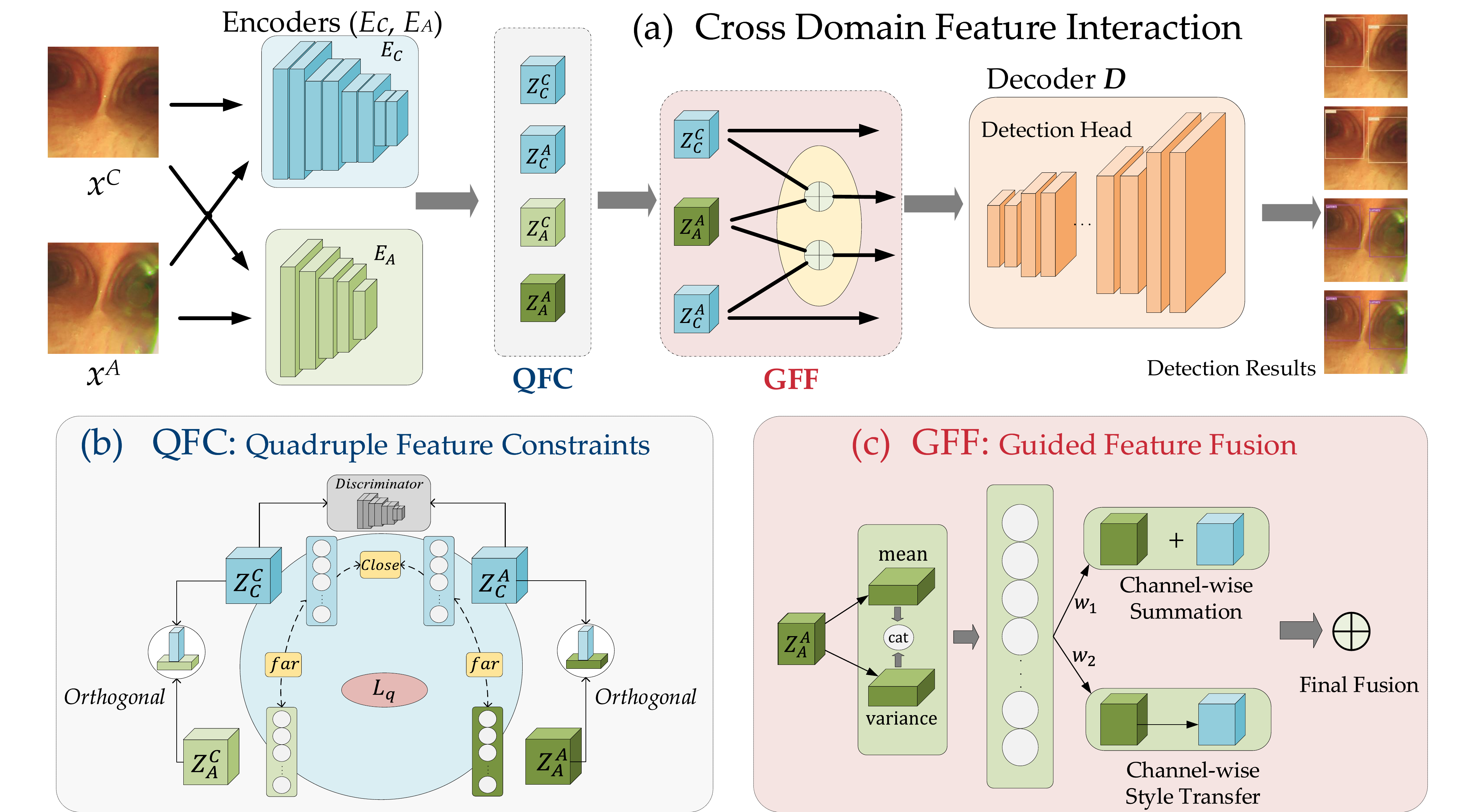}}
  \caption{Framework of the proposed Cross Domain Feature Interaction (CDFI) method. (a) presents the pipeline of our method. Images from clean and artifact domains are the input of the network. Four features are extracted from two different encoders $E_C$ and $E_A$, which later are fused to generate four detection results by the shared decoder. (b) illustrates the Quadruple Feature Constraints (QFC) module, which restricts four extracted features by the feature and distance relationships. (c) illustrates the Guided Feature Fusion (GFF) module. The mean and variance of feature $Z_A^A$ are combined to decide the weights of two different feature fusion methods.}
  \label{fig::model}
  \end{figure*}

A natural solution to this problem is domain adaptation. The knowledge of lumen structures are adapted from the clean samples (clean domain)
to the samples with artifacts (artifact domain). Therefore, the intrinsic challenge is to learn the domain-invariant features shared
by multiple domains. For general vision tasks, Sun and Saenko~\cite{sun2016deep} aligned the second-order statistical characteristics 
of the distribution of the source domain and the target domain; Bousmalis $et\ al.$~\cite{bousmalis2016domain} introduced the similarity metrics
to learn the domain-specific and domain-invariant features, while Schroff $et\ al.$~\cite{schroff2015facenet} applied the triplet loss to learn the 
discriminative features. For medical tasks, Zhang $et\ al.$~\cite{zhang2021fda} also learned the invariant
features with the domain alignment metrics. These methods can effectively associate multiple domains for knowledge generalization. 
However, they focus on the sample-based features which cannot fully exploit the intrinsic problem of lumen detection.

In this work, we propose a Cross Domain Feature Interaction (CDFI) framework for lumen detection which fully exploits
the adaptive combination of clean and artifact characteristics. The proposed method is a multi-branching 
encoder-decoder structure for feature extraction and interaction from different domains. 
To fully utilize the information of the clean and artifact features, the Quadruple Feature Constraints (QFC)
module is designed to integrally control the feature distances. Furthermore, we adapt both local and global artifacts 
and exploit the statistical characteristic of artifact features with the Guided Feature Fusion (GFF) module. The features
are finally interacted and fused to generate four output branches, which are optimized with the ground truth of lumens.
Extensive experiments demonstrate that the proposed method achieves superior accuracy on lumen detection even with strong artifacts.

\section{Method}
\label{sec:method}
We design the Cross Domain Feature Interaction (CDFI) network, 
to learn the knowledge from both clean domain and artifact domain images
 with adaptive feature fusion. The overview of our proposed method is 
 illustrated in Fig.\ref{fig::model}. 

\subsection{Cross Domain Feature Interaction Framework}\label{subsec::cdfi}

In this work, the training dataset is composed with two domains: $\{X^D, Y^D\} = \{x_i^D,y_i^D\}, i=1,2,\ldots,n^{D}, D\in\{C,A\}$ 
where $D$ is the indicator of data domain, $A$ denotes the artifact domain and $C$ denotes the clean domain; $x_i^D$ is the endoscopic images,
$y_i^D$ is the ground truth of lumen, and $n^D$ is the number of samples.

During the training, a pair of samples from both clean and artifact domains are simultaneously fed into 
the proposed model. To learn the domain-specific features, two fully convoltional encoders, $E_C$ and $E_A$, are designed to extract the clean and artifact features of 
the input images. The encoder $E_A$ is composed with less layers since the artifacts can be characterized with low-level features.
Specifically, we can obtain the clean and artifact features from clean and artifact input images as follows:
\begin{eqnarray}\label{eq::features}
  \begin{aligned}
  &z_C^C = E_C(x^C),z_A^C = E_A(x^C),\\
  &z_C^A = E_C(x^A),z_A^A = E_A(x^A).
  \end{aligned}
\end{eqnarray}
where $z_S^{S'}, S\in\{C,A\},S'\in\{C,A\}$ denotes the feature of sample $x^{S'}$ extracted by the encoder $E_{S}$.
It is expected that the encoder $E_C$ can extract the structural features of lumen from both $x^C$ and $x^A$ while $E_A$ only
focuses on the artifact features. Therefore, we propose the Quadruple Feature Constraints module in Section~\ref{subsec::qfc} to constrain
the relationship of features extracted in Eq.\eqref{eq::features}.

Since the number of training samples with artifacts is limited, training with only these samples can easily lead to the overfitting problems.
In this work, the features from $E_A$ and $E_C$ can be further combined to generate new patterns. Given a specific fusion operator $\oplus$,
the artifact feature  $z_A^A$ can be fused with $z_C^A$ and $z_C^C$ where $z_C^A\oplus z_A^A$ reconstructs the original artifact sample while
$z_C^C\oplus z_A^A$ characterizes the lumen structure of clean sample $x^C$ and the artifact feature of $x^A$. To exploit the features of artifacts,
we proposed the Guided Feature Fusion (GFF) module in Section~\ref{subsec::gff}.

Finally, these features are fed into the detection head $D$ to obtain the detection results. Four branches are designed to generate the lumen detection results from different feature 
combinations as follows:
\begin{eqnarray}
  \begin{aligned}
    &\hat{y}^C = D(z_C^C), {\hat{y}}^{CA} = D(z_C^C\oplus z_A^A), \\
    &\hat{y}^A = D(z_C^A), {\hat{y}}^{AA} = D(z_C^A\oplus z_A^A).
  \end{aligned}
\end{eqnarray}

\subsection{Quadruple Feature Constraints}~\label{subsec::qfc}
As mentioned above, the encoders $E_A$ and $E_C$ are designed to extract different types of features. In this work, the explicit constraints are proposed to guarantee that 
the features from encoders are distinguishable. Given the features in Eq.\eqref{eq::features}, we comprehensively consider the relationship and design the 
Quadruple Feature Constraints (QFC) module. It is expected that the clean features from clean and artifact domain images both characterize the structure knowledge of the lumen, 
which requires the high similarity between $z_C^C$ and $z_C^A$. Therefore, we introduce a discriminator~\cite{dann} in this module, which cannot distinguish $z_A^C$ and $z_C^C$. 
Meanwhile, the clean and artifact features of the same input image need to be different. To build a strong restriction, we consider the orthogonal relationship~\cite{yang2021dolg} to differentiate the features, 
constraining the projection from artifact features to clean features. These feature constraints can be summarized as Eq.\eqref{eq::lf}. 

\begin{equation}\label{eq::lf}
\begin{aligned}
L_f(\tilde{z}) = f_{proj}(z_A^C, z_C^C) + f_{proj}(z_A^A, z_C^A) 
+ \\\sum_{z\in{{z_C^C,z_C^A}}}\left\{ d_z\log D(z)+\left(1-d_z\right) \log \left(1-D(z)\right)\right\}
\end{aligned}
\end{equation}
where $\tilde{z}=\{z_C^C,z_A^C,z_C^A,z_A^A\}$, $f_{proj}(x, y)$ is the projection from $x$ to $y$, 
$d_z$ is the domain label of feature $z$ and $D(z)$ is the domain prediction of the discriminator.

In addition to pairwise constraints in Eq.\eqref{eq::lf}, we also consider the intrinsic connections of these features.
Inspired by triplet loss~\cite{schroff2015facenet}, the distance of the features from same domains should be close 
while the cross domain distance is larger. Since the artifact varies in location and style, we apply 
fully connected layers for each feature to eliminate this effect. The distance constraint is presented in Eq.\eqref{eq::ld} :

\begin{equation}\label{eq::ld}
\begin{aligned}
L_d(\tilde{z}) = & max(d_{fc}(z_C^C, z_C^A) - d_{fc}(z_C^C, z_A^C) + c, 0) \\ 
&+ max(d_{fc}(z_C^C, z_C^A) -d_{fc}(z_C^A, z_A^A)+ c, 0)
\end{aligned}
\end{equation}
where $d_{fc}$ represents the $L_2$ loss after fully connected layers and $c$ is the margin constant. 

As demonstrated in Fig.\ref{fig::model}(b), the feature constraint $L_f$ is illustrated by the solid lines outside the circle 
while distance constraint $L_d$ follows the  dashed lines inside the circle. We finally obtain the quadruple feature constraint loss in Eq.\eqref{eq::lq}

\begin{equation}
\begin{aligned}
L_q = L_f(\tilde{z}) + L_d(\tilde{z})
\label{eq::lq}
\end{aligned}
\end{equation}

\subsection{Guided Feature Fusion}\label{subsec::gff}
In this section, we present the design of fusion operator $\oplus$. An intuitive way is to directly add or concatenate two features. 
However, the simple summation ignores the location and type information of artifacts. 
Considering the structures and locations of the artifacts, we mainly focus on the local artifacts, which appear as the small patterns (e.g. bubbles, rainbow-like regions), and global artifacts,
which appear as the changes of global styles compared with the clean cases. 
For local artifacts, simple summation can fuse the artifact and clean features, while for global artifacts we attempt to transfer the style of artifact to the clean feature. 
In this process, we measure the global statistics of artifact features 
$z_A^A\in\mathbf{R}^{C\times H\times W}$ to capture the global artifacts as shown in Eq.\eqref{eq::stats}.
\begin{eqnarray}\label{eq::stats}
  \begin{aligned}
  &\mu(z_A^A) = \sum_{i=1}^{H}\sum_{j=1}^{W}z_A^A(i,j);\\
  &\sigma(z_A^A) = \sum_{i=1}^{H}\sum_{j=1}^{W}(z_A^A(i,j)-\mu(z_A^A) )^2
  \end{aligned}
\end{eqnarray}
where $\mu$ and $\sigma$ represents the mean and variance of the feature. Since the type and location information can be implicitly presented in the artifact features from the artifact encoder $E_A$,
we therefore expect the statistical characteristics of the extracted artifact features to determine the local and global properties of artifacts. These statistics are concatenated as follows:
\begin{equation}
C(z_A^A) = Cat(\mu(z_A^A), \sigma(z_A^A))
\end{equation}
We use $C(z_A^A)$ as weight to guide the feature fusion as follows:

\begin{eqnarray}\label{eq::gff}
  \begin{aligned}
&z_C^C\oplus z_A^A = C(z_A^A) \mathrm{AdaIN}(z_C^C, z_A^A) + (1 - C(z_A^A)) z_A^A \\
&z_C^A\oplus z_A^A = C(z_A^A) \mathrm{AdaIN}(z_C^A, z_A^A) + (1 - C(z_A^A)) z_A^A 
\label{eq3}
  \end{aligned}
\end{eqnarray}
where $\mathrm{AdaIN}(x,y)$ is the style transfer operator~\cite{huang2017arbitrary} which performs as follows:
\begin{equation}\label{eq::adain}
  \mathrm{AdaIN}(x,y) =\sigma(y)\left(\frac{x-\mu(x)}{\sigma(x)}\right)+\mu(y)
\end{equation}
Based on Eq.\eqref{eq::gff}, the fusion of features is guided by the statistics of $z_A^A$ which can simultaneously consider the global and local artifacts.
The fused features are then fed into the detection head to generate the detection results.

\begin{table*}[htbp]
\centering
\caption{The detection accuracy on Artifact Domain and 
Two Domain Test Set with different methods. The best results are in \textbf{BOLD}.}
\label{tab::main}
\begin{tabular}{c|ccc|ccc}
\hline
\multirow{2}*{Method} &\multicolumn{3}{c|}{Artifacts Domain Test Set}&\multicolumn{3}{c}{Two Domain Test Set} \\
\cline{2-7}
& \makebox[0.06\textwidth][c]{$\mathrm{AP}$} & \makebox[0.06\textwidth][c]{$\mathrm{AP_{0.5}}$} & \makebox[0.06\textwidth][c]{$\mathrm{AP_{0.75}}$}&\makebox[0.06\textwidth][c]{$\mathrm{AP}$} & \makebox[0.06\textwidth][c]{$\mathrm{AP_{0.5}}$} & \makebox[0.06\textwidth][c]{$\mathrm{AP_{0.75}}$} \\
\hline Train on CDD & 44.8 & 80.4 & 36.2 &41.8 &78.9&29.6 \\
Train on CDD, Finetune on ADD & 46.4 & 82.8& 36.8&38.5&74.9&26.0 \\
 Train on TDD & 47.7& 84.5 & 38.6&42.5&82.5&30.0 \\
\hline
 AdaIN~\cite{huang2017arbitrary} &47.7&85.1&38.6&42.9&82.0&30.4\\
CORAL Loss~\cite{sun2016deep}&48.0&85.1&39.7&43.1&82.2&30.5 \\
 Triplet Loss~\cite{schroff2015facenet} & 48.0 & 85.4& 39.3&42.5&80.9&30.5 \\
  MMD Loss~\cite{mmd} & 48.3 & 85.3& 39.2&42.7&81.1&31.3 \\
  DANN~\cite{dann} & 47.1 & 85.2 & 37.3&43.4&82.0&31.4 \\
\hline CDFI (ours) & \textbf{50.0}&\textbf{86.7}&\textbf{42.0}&\textbf{46.0}&\textbf{84.0}&\textbf{35.0} \\
\hline
\end{tabular}
\end{table*}

\subsection{Cross Domain Detection Loss}
The cross domain structure of CDFI can learn both clean and artifact domain knowledge and generate compound detection masks.
 With the lumen labels in both domains, the detection loss of YOLOv3~\cite{redmon2018yolov3} network is applied for all the detection results: 
 \begin{eqnarray}
    \begin{aligned}
 &L_C = L(\hat{y}^C, {y}^C), L_{CA} = L(\hat{y}^{CA}, {y}^C),\\
  &L_A = L(\hat{y}^A, {y}^A), L_{AA} = L(\hat{y}^{AA}, {y}^{A}).
    \end{aligned}
 \end{eqnarray}
Finally, the overall lumen detection loss can be summarized as Eq.\eqref{eq2}, where $w_\cdot$  refer to the corresponding weights for each loss.

\begin{equation}
L = w_1  L_C + w_2 L_{CA} + w_3  L_A + w_4 L_{AA} + w_q  L_q
\label{eq2}
\end{equation}

\begin{figure*}[htbp]
  \centering
  \centerline{\includegraphics[width=1.0\linewidth]{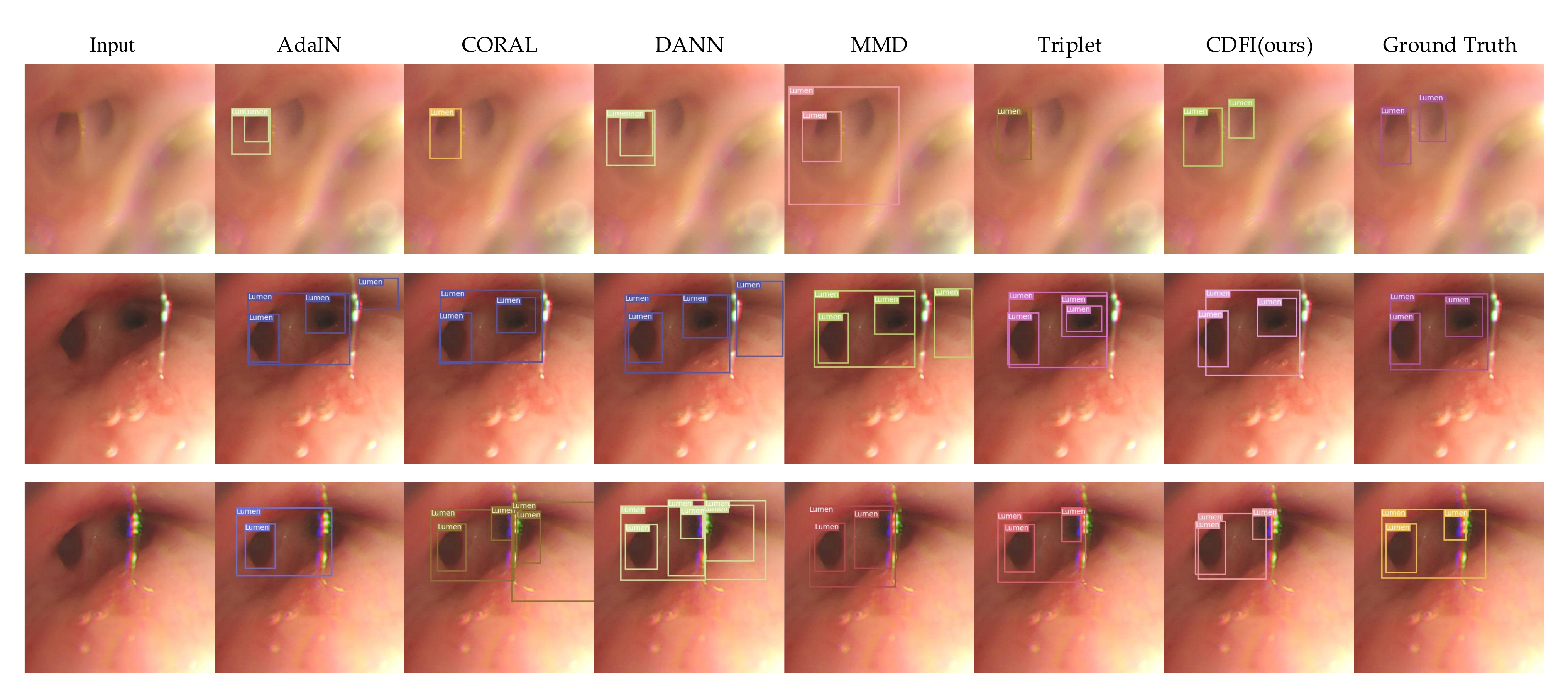}}
\caption{Qualitative results of lumen detection. Detection results of five domain adaption methods and our proposed model are presented along with the ground truth. 
Top row shows the global artifact which blurs the whole image. Middle row illustrates local artifact beside the lumen. Bottom row illustrates local artifact that blocks the lumen.}
\label{v1}
\end{figure*}

\section{Experiments and Results}
\label{sec:experiment}
\subsection{Dataset and Implementation Details}
\noindent\textbf{Dataset:} 
In the experiment, both phantom and $\textit{in-vivo}$ datasets are used to evaluate the performance of the proposed method where the images are captured with the Olympus BF-P290 bronchoscope.
 For fair comparisons, three datasets are built as follows: 
 1) Clean Domain Dataset (CDD): It contains 3818 frames from the phantom datasets and the $\textit{in-vivo}$ datasets. 
 2) Artifact Domain Dataset (ADD): It contains 2871 frames from the $\textit{in-vivo}$ datasets. 
 3) Two Domain Dataset (TDD): It contains all frames from both the clean and artifact domain datasets, which consists of 6689 images. 
 Each dataset is divided into training, validation and testing set at a ratio of 3:1:1 according to time series. 

\noindent
\textbf{Implementation Details:} In the experiments, we resize the input images to 416$\times$416. 
Random horizontal flipping and image size scaling are applied for data augmentation. During the training phase, the Adam optimizer is adopted to all the models. 
In the baseline network, the learning rate is set as 5e-4 while a smaller learning rate 3e-5 is applied for fine-tuning. 
For the proposed method, the learning rate of clean encoder $E_C$, artifact encoder $E_A$ and decoder $D$ are set as 5e-4. 
For the QFC module, the learning rate is set as 3e-6 and the constant margin $c$ is set to 100. 
Meanwhile, the learning rate for the GFF module is set as 3e-6. Specifically, the coefficient of each loss is empirically set to obtain the best performance on validation set where
 $w_1 = 2, w_2 = 2, w_3 = 2, w_4 = 2, w_q = 10$. The models are trained for 120 epochs with the batch size of 4. 
 Our proposed method and all baseline methods are implemented in Python3.9, PyTorch 1.10 with a single NVIDIA GeForce RTX 3090.

\begin{table*}[!t]
\centering
\caption{Results of the ablation study. The best results are in \textbf{BOLD}.}
\label{t2}
\begin{tabular}{c|ccc|ccc}
\hline
\multirow{2}*{Method} &\multicolumn{3}{c|}{Artifacts Domain Test Set}&\multicolumn{3}{c}{Two Domain Test Set} \\
\cline{2-7}
& \makebox[0.06\textwidth][c]{$\mathrm{AP}$} & \makebox[0.06\textwidth][c]{$\mathrm{AP_{0.5}}$} & \makebox[0.06\textwidth][c]{$\mathrm{AP_{0.75}}$}&\makebox[0.06\textwidth][c]{$\mathrm{AP}$} & \makebox[0.06\textwidth][c]{$\mathrm{AP_{0.5}}$} & \makebox[0.06\textwidth][c]{$\mathrm{AP_{0.75}}$} \\
\hline Baseline & 47.7& 84.5 & 38.6&42.5&82.5&30.0 \\
 CDFI w/ QFC only & 48.7& 86.6 & 39.1&42.7&81.8&30.0 \\
CDFI w/ GFF  only& 49.8 & \textbf{87.5}& 41.0&43.7&82.5&31.0 \\
CDFI (proposed) & \textbf{50.0}&86.7&\textbf{42.0}&\textbf{46.0}&\textbf{84.0}&\textbf{35.0} \\
\hline
\end{tabular}
\end{table*}
\begin{figure*}[!t]
  \centering
  \centerline{\includegraphics[width=1.0\linewidth]{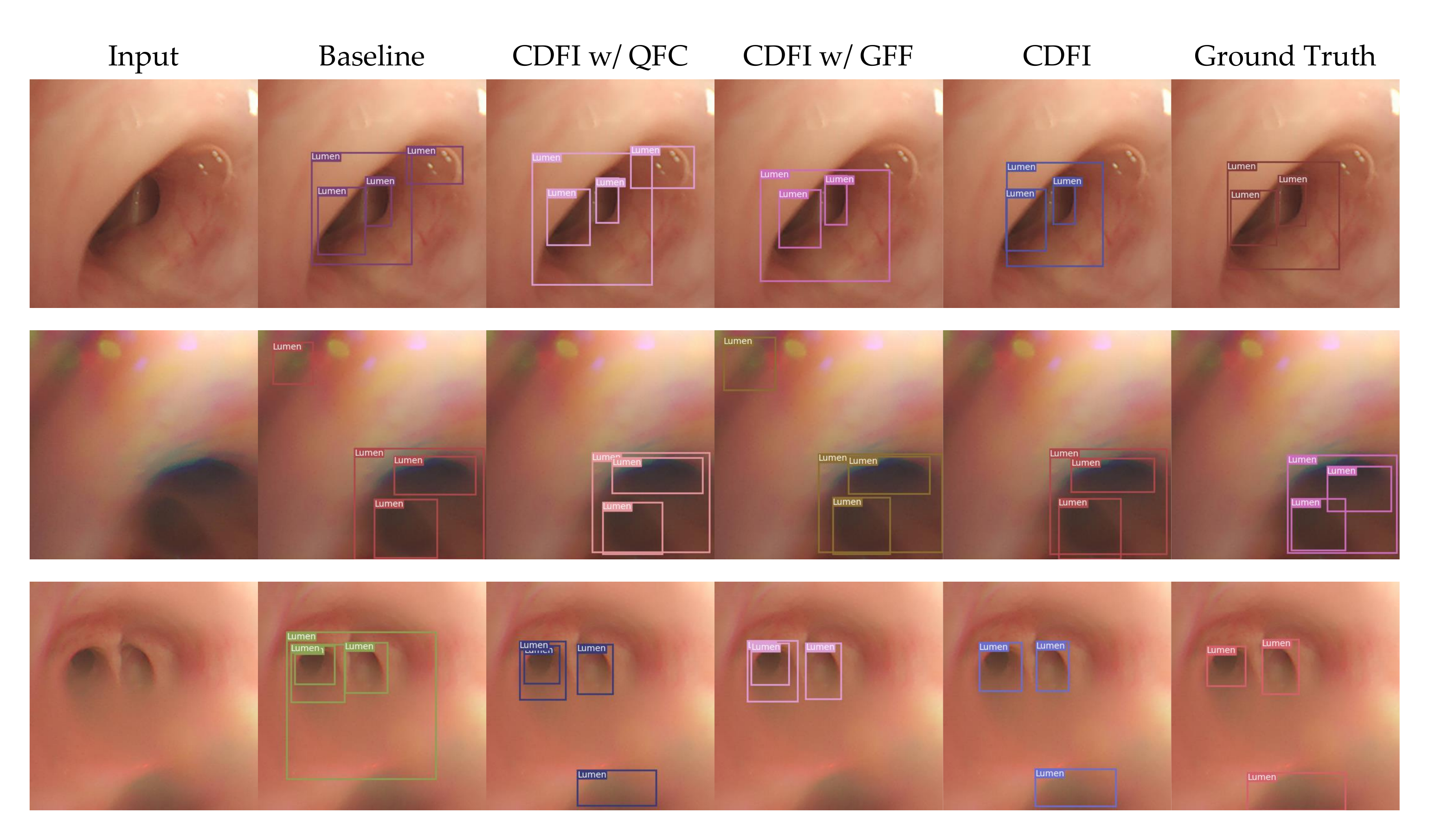}}
\caption{Visualization of ablation study. Top row shows the local bubble beside lumen. Middle row illustrates global blur and dispersion. Bottom row illustrates another global blur.}
\label{v2}
\end{figure*}


\subsection{Evaluation and Results}

\noindent
\textbf{Quantitative Results:} To evaluate the accuracy of lumen detection,
we adopt the Average Precision (AP) metric which is widely used in object detection. Different thresholds
of intersection-over-union (IoU) are used including $\mathrm{AP}$ with the full range, $\mathrm{AP_{0.5}}$ with IoU > 0.5, and $\mathrm{AP_{0.75}}$ with IoU > 0.75. All methods are first evaluated on  artifact test set. 
With the consideration of clinical application, we also test models on the images 
from both clean and artifact domains to evaluate whether the models can maintain overall 
accuracy of lumen detection. 

In the experiment, Darknet-53~\cite{redmon2018yolov3} is adopted as the network backbone. 
The simple baseline is to train the model with the standard encoder-decoder architecture 
as YOLOv3 with different setups of training data. We also implemented domain adaptation
methods including AdaIN~\cite{huang2017arbitrary}, CORAL~\cite{sun2016deep}, Triplet Loss~\cite{schroff2015facenet},
 MMD~\cite{mmd} and DANN~\cite{dann} for comparisons.

As shown in Table \ref{tab::main}, the proposed method achieves the best performance in 
$\mathrm{AP}$, $\mathrm{AP_{0.5}}$ and $\mathrm{AP_{0.75}}$ on all the test sets.
For the baseline method in the first part of the table, models finetuned with artifact images 
 can improve the accuracy on artifact test set while affect the overall performance.
This is attribute to the overfitting of artifact data.  Models trained on TDD have limited 
performance improvement since this strategy cannot fully exploit the clean and artifact features
from two domains. 

The domain adaption methods based on the dual encoders can obtain the improvement as shown in the middle part of the table. 
  AdaIN~\cite{huang2017arbitrary} transfers the style of artifact domain  to the content of clean domain while the extracted features are not restricted. 
  CORAL Loss~\cite{sun2016deep} aims to align the second-order statistical characteristics 
  of the two domains. Triplet Loss~\cite{schroff2015facenet} considers the feature 
  distance among domains to differentiate the features. MMD Loss~\cite{mmd} finds 
  the smallest  moment of order among clean and artifact domains and 
  achieves a better detection results on artifact data. 
  DANN~\cite{dann} method introduces Gradient Reversal Layer (GRL) to 
  confuse the domain discriminator and attains an increase in overall detection. 

  To learn the knowledge of both clean and artifact domains, we design two modules to
   accomplish feature Constraints and adaptive feature fusion. The QFC module integrates 
   the features to learn the complementary encoders. The GFF module  utilizes the statistical 
   characteristics of artifact features, 
   realizing appropriate feature fusion. The results demonstrate that the proposed method
    achieves  superior lumen detection results on artifacts images and, more importantly, can
    preserve the best accuracy on both clean and artifact domains.

\noindent
\textbf{Qualitative Results:} Fig.\ref{v1} presents the qualitative results of different methods. 
This figure gives the lumen detection results of all the dual input methods
 on three artifact images. 
 The first row presents the global artifact which blurs the whole image. 
 The domain adaption methods fail to detect the small lumen on the right due to
  the severe image blur. 
 The second and third rows illustrate the local artifact caused by the reflection.
  The artifact beside the lumen can be characterized as the boundary of lumen
  which leads to incorrect detection mask, as shown in the second row. 
In the bottom row, the lumen occluded by the artifact is hard to detect.
   Furthermore, false masks will be generated around the artifacts. Compared with other methods,
   the proposed method achieves successful detection results in these cases.

\subsection{Ablation Study}
We conduct the ablation study to measure the impact of each module 
in the proposed method. 
The model trained on TDD is set as the baseline. 
Since GFF module has no constraints to the encoders, 
we add the discriminator and orthogonal operator to constrain 
the extracted features before feature fusion when evaluating the  GFF module. 
Then we test CDFI with QFC module and CDFI with GFF module separately.


TABLE \ref{t2} presents the results of the ablation study. It can be observed
 that QFC module improves the accuracy of the artifact images, 
 which means the inner relationship among extracted features are well modeled.
  QFC module ensures that two encoders successfully extract clean and 
  artifact features. GFF module achieves a great improvement on the 
  artifact images, which is attribute to the adaptive feature fusion process. 
  GFF module effectively utilizes the artifact information and distribute 
  appropriate weights for feature fusion to have a great perception to 
  lumen with strong artifacts.

 The combination of the two modules achieves a significant improvement on TDD test set, 
 which proves the effectiveness of the CDFI method. 
 Robust feature constraints from QFC and reliable feature fusion from GFF 
 are combined to deliver stable lumen detection results.

  Qualitative results in Fig.\ref{v2} demonstrate the effectiveness 
  of each module. The top row illustrates the local artifact, 
  a bubble, beside the lumen. Baseline and QFC methods tend to 
  recognize the bubble as lumen while GFF module attains correct detection.
   The middle row shows global blur and dispersion, which confuse the 
   baseline and GFF methods to recognize the artifact at top left 
   corner as lumen. QFC module performs well in this case. 
   The bottom row illustrates another global image blur, 
   which leads to missing detection and incorrect detection for 
   most of the methods while CDFI method outputs the correct detection. 
   The results show that our designed modules can achieve great performance 
   in artifact domain images. Meanwhile, the combination of the modules can 
   achieve further improvement for lumen detection tasks.

\section{Conclusion}
In this work, we propose a CDFI network to learn the intrinsic artefact model and achieve great lumen detection results. 
To address the problems caused by artifacts, we design a QFC module to integrate and regularize the feature information 
from both clean domain and artifact domain images. In addition, a GFF module is designed to realize adaptive feature fusion 
instructed by the statistical characteristics of the artifact features. Extensive experiments demonstrate that our model achieves superior lumen 
detection results compared to current state-of-the-art, which is beneficial for the navigation of surgical robots in bronchi.


\end{document}